\documentclass[10pt,twocolumn,letterpaper]{article}

\usepackage[pagenumbers]{cvpr} 

\definecolor{cvprblue}{rgb}{0.21,0.49,0.74}
\usepackage[pagebackref,breaklinks,colorlinks,allcolors=cvprblue]{hyperref}

\def\paperID{15356} 
\def\confName{CVPR}
\def\confYear{2026}

\title{CircleFlow: Flow-Guided Camera Blur Estimation using a Circle Grid Target}

\author{Jiajian He$^1$ \quad Enjie Hu$^1$ \quad Shiqi Chen$^2$ \quad Tianchen Qiu$^1$ \quad Huajun Feng$^1$ \quad Zhihai Xu$^1$ \quad Yueting Chen$^{1\dag} $\\
$^1$Zhejiang University \quad $^2$University of California, Los Angeles
}

\begin{document}
\twocolumn[{%
\renewcommand\twocolumn[1][]{#1}%
\maketitle
\begin{center}
\vspace{-0em}
    \centering
    \captionsetup{type=figure}
   \includegraphics[width=\linewidth]{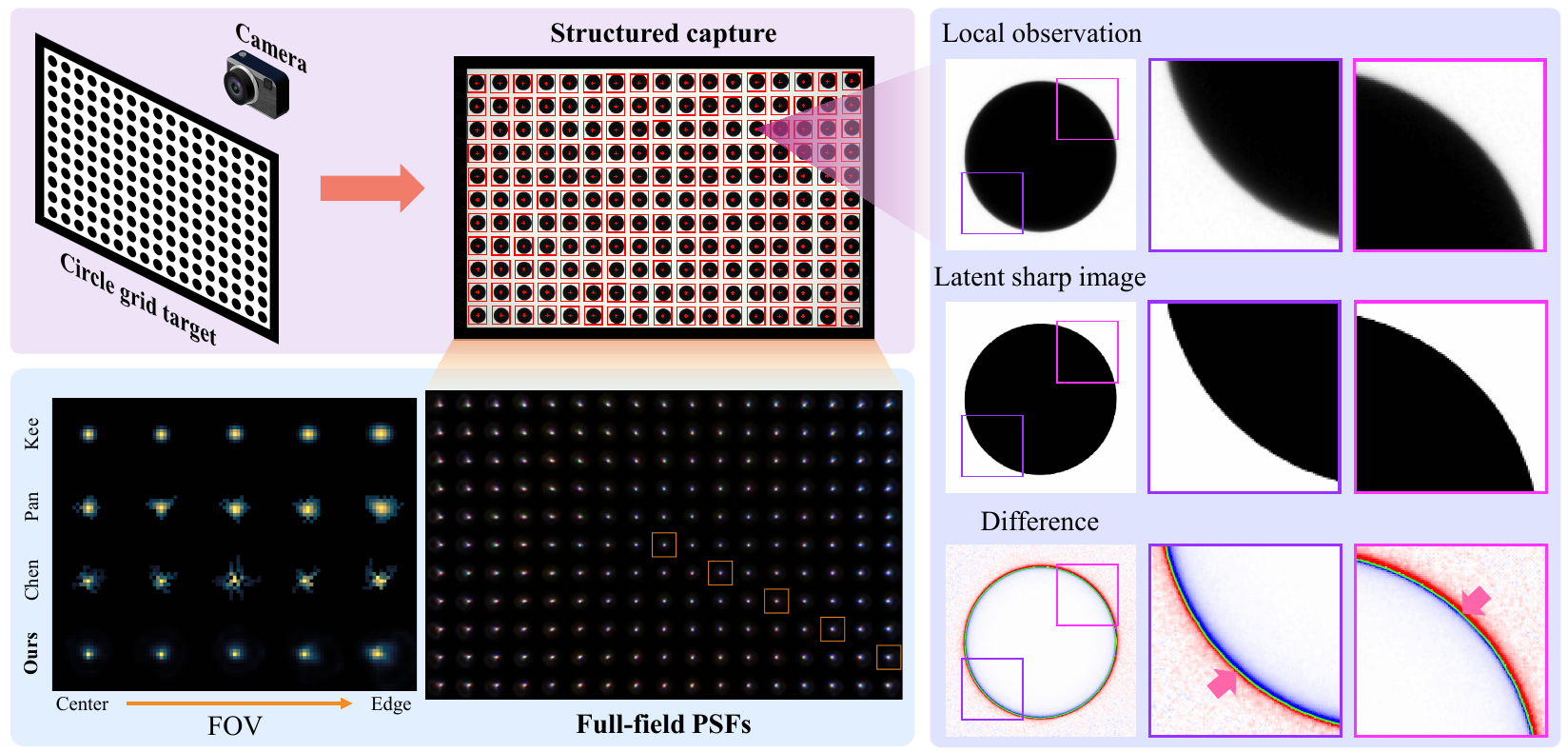}
    \caption{
    We introduce CircleFlow, a high-fidelity PSF estimation framework that employs flow-guided edge localization for precise blur characterization. 
    Top-left: a circle grid chart is captured to provide structured observations of field-dependent blur. 
    Right: for a selected patch, the local blurred observation, the reconstructed latent sharp image, and their difference are shown, with two zoomed-in regions highlighting sub-pixel edge alignment (green) and the residual blur with noise visualized in red/blue. 
    Bottom-right: full-field PSF calibration results. 
    Bottom-left: comparison of PSFs (green channel) from different methods at five field positions from center to edge.
    }
   \label{fig:teaser}
   \vspace{-0em}
\end{center}
}]

\let\thefootnote\relax\footnotetext{$^{\dag}$ Corresponding author.}

\begin{abstract}
The point spread function (PSF) serves as a fundamental descriptor linking the real-world scene to the captured signal, manifesting as camera blur. Accurate PSF estimation is crucial for both optical characterization and computational vision, yet remains challenging due to the inherent ambiguity and the ill-posed nature of intensity-based deconvolution. We introduce CircleFlow, a high-fidelity PSF estimation framework that employs flow-guided edge localization for precise blur characterization. CircleFlow begins with a structured capture that encodes locally anisotropic and spatially varying PSFs by imaging a circle grid target, while leveraging the target’s binary luminance prior to decouple image and kernel estimation. The latent sharp image is then reconstructed through subpixel alignment of an initialized binary structure guided by optical flow, whereas the PSF is modeled as an energy-constrained implicit neural representation. Both components are jointly optimized within a demosaicing-aware differentiable framework, ensuring physically consistent and robust PSF estimation enabled by accurate edge localization. Extensive experiments on simulated and real-world data demonstrate that CircleFlow achieves state-of-the-art accuracy and reliability, validating its effectiveness for practical PSF calibration.
\end{abstract}
    
\section{Introduction}

The point spread function (PSF) describes how a camera maps scene points to sensor intensities and is a fundamental descriptor of an imaging system’s optical quality. 
In optics, the PSF encodes aberrations and guides lens design~\cite{zhou2024revealing,zhang2024end,ren2025successive}, alignment~\cite{hu2025fast,zhou2024optical}, and tolerance analysis~\cite{ren2025fast,zhou2025optical};  
in computational vision, accurate PSF estimation enables physically grounded image restoration~\cite{chen2021optical,chen2021extreme,diamond2021dirty,chen2022computational,zhang2021plug}. 
Reliable PSF estimation is therefore essential for both optical characterization and computational vision.

Directly estimating the PSF from a single blurred observation is intrinsically difficult.   
The observed blur results from the convolution of an unknown sharp image and an unknown kernel, and multiple combinations of these two can reproduce the same observation~\cite{li2023self}. 
Without strong priors, optimization often collapses to trivial or unstable solutions.
A practical remedy is target-based PSF calibration, where a structured pattern with known geometry constrains the solution space.  
Checkerboards~\cite{chen2021extreme,chen2025physics} provide sharp edges but only along two orientations, limiting their ability to encode directional blur.  
Noise and texture-based patterns~\cite{delbracio2012non,lin2025learning,mosleh2015camera} offer broadband content but require affine pre-registration, where small localization errors may degrade kernel accuracy.  
Thus, achieving both complete and precise characterization of field-dependent and directionally varying blur remains challenging.

A key insight in blur estimation is the importance of edge localization~\cite{joshi2008psf,pan2016l_0,fergus2006removing,cho2009fast,xu2013unnatural,krishnan2011blind}, since blur manifests most distinctly around edges where its spatial spread encodes the underlying PSF.
\textit{Accurate edge localization directly determines the quality of the recovered PSF.}

Motivated by this insight, we propose \textbf{CircleFlow}, a high-fidelity PSF estimation framework that exploits structured capture and flow-guided edge localization, as shown in~\cref{fig:teaser}.
Our approach begins with imaging a circle grid target, whose continuous contours provide complete directional gradient responses for encoding spatially varying PSFs with locally anisotropic structures, while its binary luminance structure supplies a strong prior that helps decouple image and kernel estimation.
Based on this structured observation, CircleFlow reconstructs the latent sharp image through subpixel optical-flow alignment of an initialized binary structure and models the PSF using an energy-constrained implicit neural representation (INR)~\cite{sitzmann2020implicit}. 
Both components are jointly optimized within a demosaicing-aware differentiable framework, yielding robust and physically faithful PSF estimation.

Our main contributions are as follows:
\begin{itemize}
\item We leverage a circle grid target whose circular contours provide continuous directional gradients and redundant boundary sampling, yielding complete and robust encoding of spatially varying and locally anisotropic blur.
\item We introduce a flow-guided refinement module that uses the binary luminance prior to achieve subpixel edge localization and resolve ambiguity between structure and blur.
\item We present CircleFlow as a reliable PSF estimation framework for low-level vision, recovering real aberration-induced blur and supporting downstream tasks like image restoration and optical-quality evaluation.
\end{itemize}

We validate CircleFlow through extensive simulation and real-world experiments.
As illustrated in Fig.~\ref{fig:teaser}, the method reconstructs sharp edges and accurately recovers spatially varying PSFs across the field.
The reconstructed PSFs closely match ray-traced ground truth in simulation (\cref{fig:simulation results}), align with measured spatial frequency response (SFR) in real camera calibration (\cref{fig:sfr_measurement}), and further enables effective deblurring of real images (\cref{fig:deblur}).
These results collectively demonstrate the accuracy, robustness, and practical utility of the proposed framework.
\section{Related Work}

\paragraph{Target-Based PSF Calibration.}
Structured targets have long been used for camera calibration and blur characterization due to their controllability and repeatability~\cite{brauers2010direct,zhang1999flexible,chen2021extreme,chen2025physics,delbracio2012non,mosleh2015camera,lin2025learning}.  
Checkerboard patterns are widely adopted for geometric calibration~\cite{zhang1999flexible,labussiere2022leveraging} and have been extended to estimate PSFs from edge spread functions~\cite{chen2021extreme,chen2025physics}.  
However, their edges are aligned only along two orthogonal directions, providing incomplete blur representation.  
Noise- and texture-based targets~\cite{delbracio2012non,mosleh2015camera,lin2025learning} offer broadband frequency coverage but rely on affine or homographic registration, where localization errors can introduce misalignment.  
Our method uses a circle-grid target with a binary luminance prior to improve calibration accuracy and robustness without requiring affine pre-alignment.

\paragraph{PSF Modeling.}
PSF modeling can be broadly divided into parametric~\cite{markham1999parametric,schuler2012blind,kee2011modeling,eboli2022fast} and non-parametric~\cite{kee2011modeling,joshi2008psf,gwak2020modeling,cho2011blur} approaches.  
Parametric models, such as Gaussian mixtures~\cite{kee2011modeling,eboli2022fast} or polynomial expansions~\cite{niu2022zernike,chen2025physics}, represent the blur kernel with a small number of interpretable parameters, offering simplicity but limited flexibility for complex aberrations.  
Non-parametric models provide greater flexibility but often overfit and become unstable under noise or low signal-to-noise conditions.  
They require strong regularization to maintain plausible solutions and remain sensitive to initialization with limited generalization.  
Our approach employs INR~\cite{sitzmann2020implicit,shen2022nerp,lin2022non} with energy-constrained optimization and structured priors to achieve coherent and interpretable PSF representations.

\paragraph{PSF Estimation.}
A large body of research has studied PSF estimation as an inverse problem~\cite{savakis1993blur,babacan2008variational,shan2008high,xu2010two,pan2016l_0,kee2011modeling,mosleh2015camera,eboli2022fast,pan2014deblurring,chen2021extreme,li2023self,chen2025physics,lin2025learning,bell2019blind,liang2021mutual,pan2019phase}.  
Classical deconvolution methods rely on handcrafted priors such as total variation (TV)~\cite{rudin1992nonlinear,babacan2008variational}, frequency-domain constraints~\cite{savakis1993blur,mosleh2015camera,pan2019phase}, or gradient prior~\cite{shan2008high,xu2010two,pan2016l_0,ChenNature2025} to suppress noise and stabilize estimation.  
While effective in simple cases, these methods are sensitive to parameter tuning~\cite{chen2021extreme}.  
Neural approaches~\cite{ulyanov2018deep,ren2020neural,li2023self,bell2019blind,lin2025learning,ji2020real} such as Deep Image Prior~\cite{ulyanov2018deep,ren2020neural,li2023self} exploit the implicit regularization of convolutional architectures~\cite{shi2022measuring} to jointly recover the latent image and kernel.  
However, without explicit constraints or structural guidance, they may converge to trivial identity solutions, especially with calibration patterns of limited texture.  
Our method reformulates PSF estimation by leveraging precise edge localization, transforming intensity-based deconvolution into a structure-aware estimation process that enhances robustness.

\section{Method}

\subsection{Overview}
\label{sec:overview}

During image formation, light rays from the scene pass through the lens system and are converted into digital signals by the image sensor.  
Within a small local region of the sensor where PSF variation is negligible, the captured image can be modeled as
\begin{equation}
    b = \mathcal{B}(i * k) + n,
    \label{eq:convolution}
\end{equation}
where $i$ is the latent sharp image, $k$ is the local PSF, $\mathcal{B}$ denotes sensor sampling, and $n$ represents measurement noise.  
Given the blurred observation $b$, PSF estimation can be formulated as a joint inverse problem under the Maximum a Posteriori (MAP) framework~\cite{levin2011efficient}:
\begin{equation}
i, k = \arg\min_{i,k} \Vert i * k - b \Vert^2 + \lambda \mathrm{P}(i) + \beta \mathrm{Q}(k),
\label{eq:least squares question}
\end{equation}
where $\mathrm{P}(i)$ and $\mathrm{Q}(k)$ impose regularization on the image and PSF, and $\lambda$, $\beta$ control their relative weights.  

This problem is inherently ill-posed because multiple $(i, k)$ pairs can reproduce the same blurred observation~\cite{li2023self}.  
Accurate edge localization is therefore essential, as blur manifests most distinctly along edges where its spatial spread encodes the underlying PSF.  
Precise alignment of these edge structures allows the coupling between geometry and blur to be effectively disentangled, forming the foundation for reliable kernel estimation.

To address this challenge, we propose CircleFlow, a high-fidelity PSF estimation framework that achieves precise blur characterization through flow-guided edge localization.  
The process begins with a structured capture using a circle grid target that encodes field-dependent blur through spatially distributed circular elements.  
The target’s inherent binary luminance pattern further introduces a strong prior, helping decouple the latent image from kernel estimation.  

\begin{figure*}[t]
    \centering
    \includegraphics[width=\linewidth]{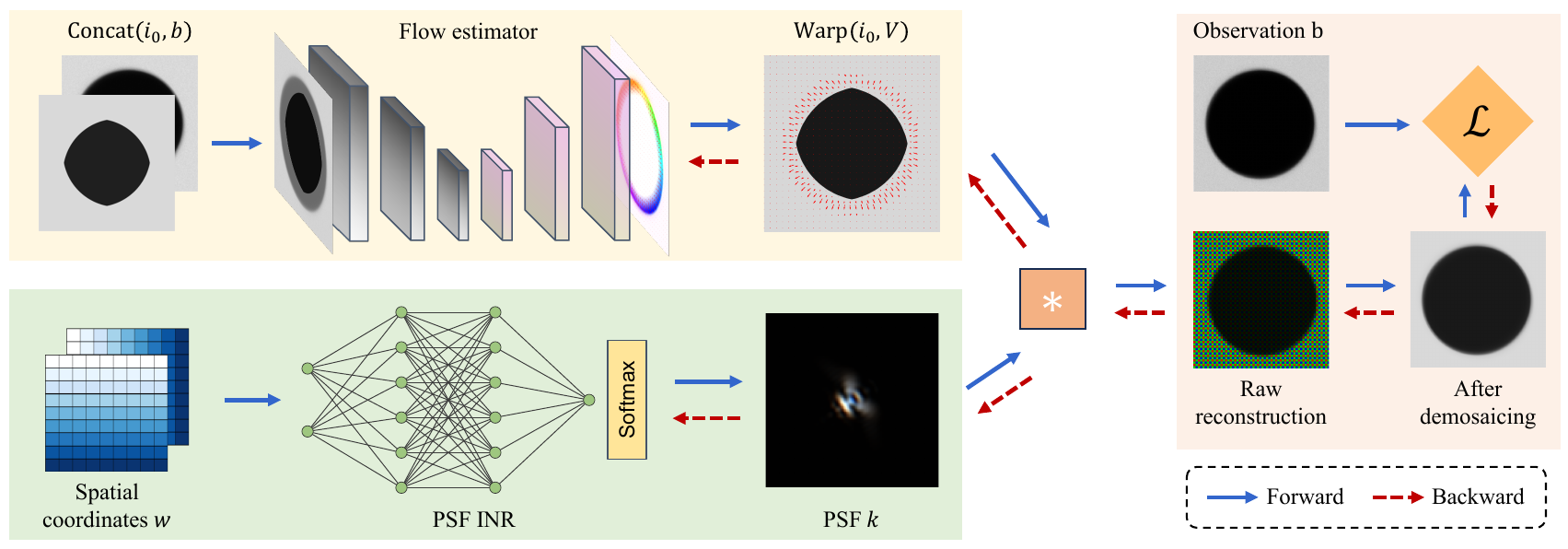}
    \caption{
    \textbf{Overview of the demosaicing-aware joint optimization framework in CircleFlow.}
    The demosaiced observation $b$ and the binary proxy $i_0$ are concatenated and fed into a flow estimator that predicts a dense, pixel-wise deformation field $V$, yielding the geometrically aligned image $\mathrm{Warp}(i_0,V)$ under the binary luminance prior encoded in $i_0$.
    In parallel, the PSF $k$ is represented by an implicit neural representation (INR) queried at spatial coordinates $w$ and normalized via softmax, enforcing the PSF’s energy constraint.
    Both components are optimized jointly within a demosaicing-aware reconstruction loop:
    the reconstructed sharp image and the estimated PSF are convolved to synthesize a re-blurred image, which is then passed through Bayer sampling and demosaicing to match the observed $b$.
    This process mitigates interpolation artifacts introduced during demosaicing and enables consistent refinement of geometric alignment and blur modeling for accurate PSF estimation.
    }
    \label{fig:recon_pipe}
\end{figure*}

Building upon this structured observation, CircleFlow reconstructs the latent sharp image through sub-pixel optical flow alignment while simultaneously modeling the PSF using an energy-constrained implicit neural representation (INR)~\cite{sitzmann2020implicit}.  
Both components are jointly optimized within a demosaicing-aware differentiable framework, leading to accurate and stable PSF estimation that remains faithful to the underlying imaging process, as illustrated in~\cref{fig:recon_pipe}.

\subsection{Structured Capture with a Circle Grid Target}
\label{sec:characterization}

We employ a circle grid target composed of uniformly arranged black circles on a bright background to provide structured observations of camera blur. This design simultaneously ensures structured blur encoding and introduces a binary luminance prior that constrains the latent sharp image during reconstruction.

\begin{figure}[t]
    \centering
    \includegraphics[width=\linewidth]{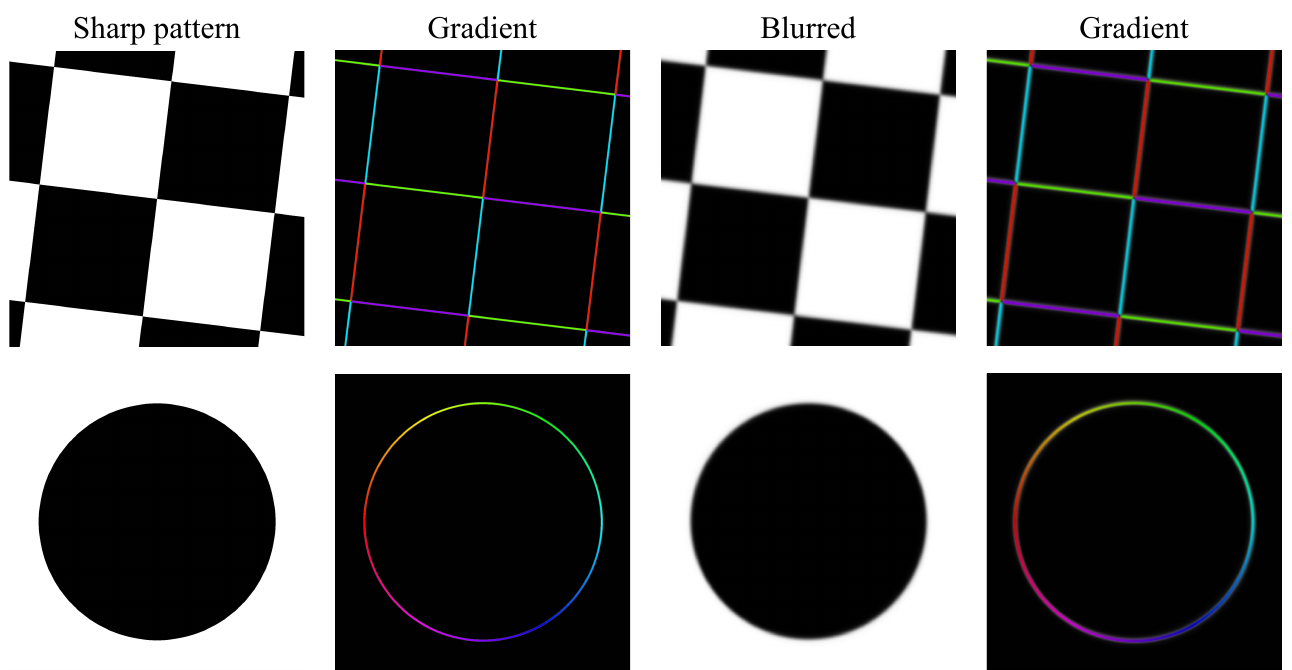}
    \caption{
    Comparison of gradient distributions for a slanted checkerboard and a circle before and after blurring.  
    Hue indicates gradient orientation, and edge width represents blur extent.  
    The checkerboard encodes blur only along two fixed directions, whereas the circle retains a continuous directional field, capturing PSF shape and directional spread more completely.
    }
    \label{fig:circle_vs_checker}
\end{figure}

\paragraph{Circular Encoding of Blur.}
In real imaging systems, the PSF varies spatially across the field of view and often exhibits locally anisotropic structures due to optical aberrations.  
To characterize such complex blur reliably, the calibration target must encode directional information that reflects how the PSF spreads across space.  
The circular boundary naturally fulfills this requirement: as a continuous curve with rotational symmetry, its local normal direction varies smoothly with angle, \textit{producing dense gradient responses along all orientations}.  
This continuous geometric structure ensures that local blur is encoded comprehensively.  
As illustrated in \cref{fig:circle_vs_checker}, the circular boundary preserves complete directional gradient coverage after blurring, whereas the checkerboard pattern retains gradients only along two orthogonal directions, limiting its ability to represent directional spread.  

Moreover, the circle’s closed geometry \textit{introduces redundant directional sampling along its edge}, effectively increasing the signal-to-noise ratio and improving robustness against noise and local variation.  
When arranged in a regular grid, the circles \textit{serve as spatially distributed probes} that jointly capture field-dependent and directionally varying PSFs from a single capture.  
Together, these structural advantages make the circle grid a compact and information-complete calibration target for full-field PSF measurement.

\paragraph{Binary Luminance Prior and Edge Guidance.}
The circle grid provides a clear binary separation between dark and bright regions, forming a natural two-level luminance prior.  
This prior confines the latent sharp image to a binary manifold, converting the recovery task into geometric alignment rather than pixel-level intensity regression, which substantially reduces the ambiguity of \cref{eq:least squares question}.  
In addition, the smooth and continuous boundary of each circle defines a stable and predictable gradient field that serves as reliable guidance for optical flow estimation, facilitating sub-pixel edge alignment during reconstruction.

\begin{figure*}[t]
    \centering
    \includegraphics[width=\linewidth]{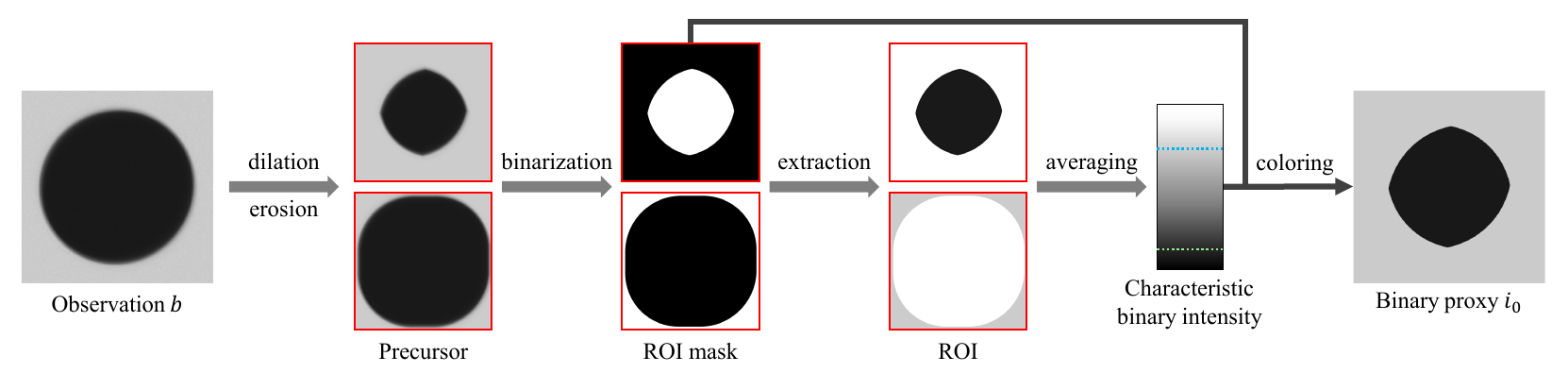}
    \caption{
    \textbf{Construction of the binary proxy $i_0$.}  
    Starting from the blurred observation $b$, morphological dilation and erosion are applied independently to suppress blur transitions in dark and bright regions.  
    The resulting precursor is binarized using Otsu’s adaptive thresholding to extract the dark and bright ROIs, whose mean intensities define the characteristic black and white levels of the latent sharp image.  
    Combining these regions yields the binary proxy $i_0$, which preserves the binary luminance pattern and serves as a reliable geometric initialization for subsequent flow-guided refinement.
    }
    \label{fig:binary_prior}
\end{figure*}

\subsection{Flow-Guided Image Reconstruction}
\label{sec:flow_reconstruction}

Directly optimizing \cref{eq:least squares question} without structural priors often leads to degenerate solutions where $i \approx b$ and $k \approx \delta$~\cite{li2023self}.  
To prevent this, CircleFlow incorporates a binary luminance prior that provides a strong structural initialization and guides the subsequent optical-flow-based geometric refinement.

\subsubsection{Binary Proxy Initialization.}
As illustrated in~\cref{fig:binary_prior}, for each local observation $b$, we first generate a precursor by applying morphological dilation and erosion as two independent operations in parallel.  
Dilation suppresses blur transitions in dark regions, while erosion suppresses them in bright regions.  
This dual filtering mitigates PSF-induced gray transitions and preserves the separability between the foreground and background.  

Subsequently, Otsu’s adaptive thresholding is applied to the precursor to produce two region-of-interest (ROI) masks corresponding to the dark and bright areas.  
These masks are used to extract the respective ROIs, and their mean intensities define the characteristic black and white levels of the latent sharp image, reducing photometric ambiguity introduced by sensor noise.  

Finally, the dark-region ROI mask is adopted as structural support, and the two intensity levels are assigned to form the binary proxy $i_0$.  
This proxy maintains the intended binary luminance pattern and provides a reliable geometric initialization for the subsequent flow-guided refinement, where sub-pixel boundary alignment is recovered jointly with PSF estimation.

\subsubsection{Flow-Guided Geometric Alignment.}
Traditional calibration methods~\cite{mosleh2015camera,lin2025learning} rely on global parametric transformations such as affine or homographic mappings derived from feature detection.  
These global models are coarse in nature and may fail to capture local variations, especially under severe noise and blur.  

To achieve robust and fine-grained edge localization, we employ \textit{optical flow for dense, pixel-wise geometric alignment}.  
Unlike global transforms, optical flow operates continuously across the image, allowing spatially adaptive correction.  
A learnable flow network, conditioned on the binary proxy $i_0$ and the blurred observation $b$, predicts a deformation field $V$ that warps $i_0$ toward the latent sharp image $i$:
\begin{equation}
V = \mathcal{G}_{\theta_i}(\mathrm{Concat}(i_0,b)), \qquad
i = \mathrm{Warp}(i_0,V),
\label{eq:Flow_Aligner}
\end{equation}
where $\mathcal{G}_{\theta_i}$ denotes the optical-flow estimator and $\mathrm{Warp}(\cdot)$ is a differentiable warping operator implemented via bilinear interpolation.   

The binary proxy provides stable geometric anchors, while the blurred observation contributes gradient cues encoding local blur characteristics.  
By leveraging these complementary cues, the flow network continuously refines geometric correspondence under the constraint of $i_0$ and learns deformation fields that remain consistent with the underlying blur process.  
This enables the model to disentangle structural misalignment from blur-induced intensity spread, ensuring accurate edge localization even under strong degradation.  

We adopt SPyNet~\cite{ranjan2017optical} as the backbone of $\mathcal{G}_{\theta_i}$ due to its hierarchical coarse-to-fine refinement and lightweight design.  
Each pyramid stage predicts residual flow updates, progressively stabilizing local deformation even in heavily blurred regions.

\subsection{Implicit Neural PSF Modeling}
\label{sec:psf_modeling}

Conventional parametric PSF models, such as Gaussian mixtures~\cite{kee2011modeling,eboli2022fast} or polynomial expansions~\cite{niu2022zernike,chen2025physics}, are constrained by fixed functional representations and simplifying assumptions, making them incapable of capturing the complex aberration characteristics exhibited in real optical systems.

The PSF fundamentally represents an \emph{energy distribution function} that describes how light from a single point in object space spreads over the image plane.  
To capture this distribution with high flexibility, we employ an INR parameterized by a multilayer perceptron (MLP)~\cite{sitzmann2020implicit}:
\begin{equation}
k = \mathcal{G}_{\theta_k}(w),
\label{eq:PSF_Solver}
\end{equation}
where $\mathcal{G}_{\theta_k}$ denotes the MLP and $w \in \mathbb{R}^2$ represents spatial coordinates within the PSF domain.  
Each coordinate query outputs a scalar response $\tilde{k}(w)$, which is subsequently normalized through a softmax function:
\begin{equation}
k(w) = \frac{\exp(\tilde{k}(w))}{\sum_{w' \in \Omega}\exp(\tilde{k}(w'))},
\label{eq:softmax}
\end{equation}
where $\Omega$ denotes the spatial support of the PSF.  

This normalization enforces two essential physical constraints that every PSF must satisfy:
\begin{equation}
k(w) \geq 0, \qquad \sum_{w \in \Omega} k(w) = 1.
\label{eq:PSF_constraints}
\end{equation}
The non-negativity ensures physically valid light intensity, while the unit-sum constraint guarantees energy conservation during image formation.  
Together, these constraints remove the trivial scaling ambiguity between the image and kernel during joint optimization.  
The softmax activation thus provides a natural and differentiable mechanism to embed energy constraints directly into the neural PSF representation, enabling stable and meaningful kernel learning within the CircleFlow framework.

\subsubsection{Demosaicing-Aware Joint Optimization}
\label{sec:joint_optimization}

Finally, the flow-guided image reconstruction and INR-based PSF modeling are jointly optimized within a differentiable framework, as illustrated in~\cref{fig:recon_pipe}.  
Conventional PSF estimation methods typically operate directly on demosaiced RGB data, where interpolation introduces non-optical spatial correlations and distorts the true PSF structure.  
In contrast, CircleFlow adopts a demosaicing-aware scheme that explicitly preserves the optical characteristics of the blur throughout optimization.  

During optimization, the optimization objective is shared between the image and kernel branches.  
At each iteration, the reconstructed image $i$ is convolved with the estimated PSF $k$ to generate a re-blurred image.  
This iterative coupling allows the two modules to refine each other: accurate geometric alignment enables more reliable PSF estimation, while improved PSF modeling provides sharper feedback for flow correction.  
The re-blurred image is then subjected to the same Bayer sampling and interpolation process as the observed image $b$, producing $\hat{b}$.  
This symmetric treatment mitigates demosaicing-induced artifacts and ensures that reconstruction remains consistent with the demosaiced observation, preserving fidelity to the sensor’s imaging process.

A fidelity loss enforces global brightness consistency:
\begin{equation}
\mathcal{L}_{fidelity} = \Vert \hat{b} - b \Vert_1,
\label{eq:fidelity_loss}
\end{equation}
while a gradient-based loss emphasizes discrepancies along edges where blur is most perceptible:
\begin{equation}
\mathcal{L}_{gradient} = \Vert \nabla_x \hat{b} - \nabla_x b \Vert_1 
                      + \Vert \nabla_y \hat{b} - \nabla_y b \Vert_1.
\label{eq:gradient_loss}
\end{equation}

The total loss integrates both terms:
\begin{equation}
\mathcal{L} = \mathcal{L}_{fidelity} + \mathcal{L}_{gradient}.
\label{eq:total_loss}
\end{equation}

Through this joint optimization, CircleFlow achieves coherent alignment between geometric structure and optical blur.  
The optical flow ensures precise edge localization, while the implicit PSF model accurately captures the energy distribution of blur, yielding reliable PSF estimation suitable for practical camera calibration
\section{Experiments and Results}
\label{sec:experiments}

We evaluate CircleFlow through both simulation and real-world experiments to verify its effectiveness and accuracy.  
The experiments focus on three aspects:  
(1) the completeness of structured blur encoding,  
(2) the effectiveness of flow-guided geometric reconstruction, and  
(3) the fidelity of the recovered PSFs compared with optical references.

For comparison, we include three representative baselines:  
\textbf{Kee et al.}~\cite{kee2011modeling} estimate PSFs via conjugate gradient descent under a Gaussian prior;  
\textbf{Pan et al.}~\cite{pan2016l_0} employ $L_0$-norm sparsity regularization, conceptually related to our binary luminance prior;  
and \textbf{Chen et al.}~\cite{chen2021extreme} use a checkerboard target combined with a deep linear model.

\subsection{Simulation Experiments}
\label{sec:simulation}

We first conduct simulation experiments using a physically-based imaging model to quantitatively assess the accuracy and robustness of CircleFlow.  
A smartphone telephoto lens is modeled by ray tracing to generate field-dependent PSFs across the full field of view.  
To simulate realistic imaging, ideal target patterns undergo random affine transformations to emulate geometric distortion.  
Each transformed pattern is convolved with the corresponding PSF to produce spatially varying blur.  
Mixed Gaussian–Poisson noise with a variance of approximately $0.01$ is added, and the images are Bayer-sampled to form synthetic RAW measurements. 
Different methods are then applied to these simulated images to estimate the PSFs.

\begin{figure}[t]
    \centering
    \includegraphics[width=0.95\linewidth]{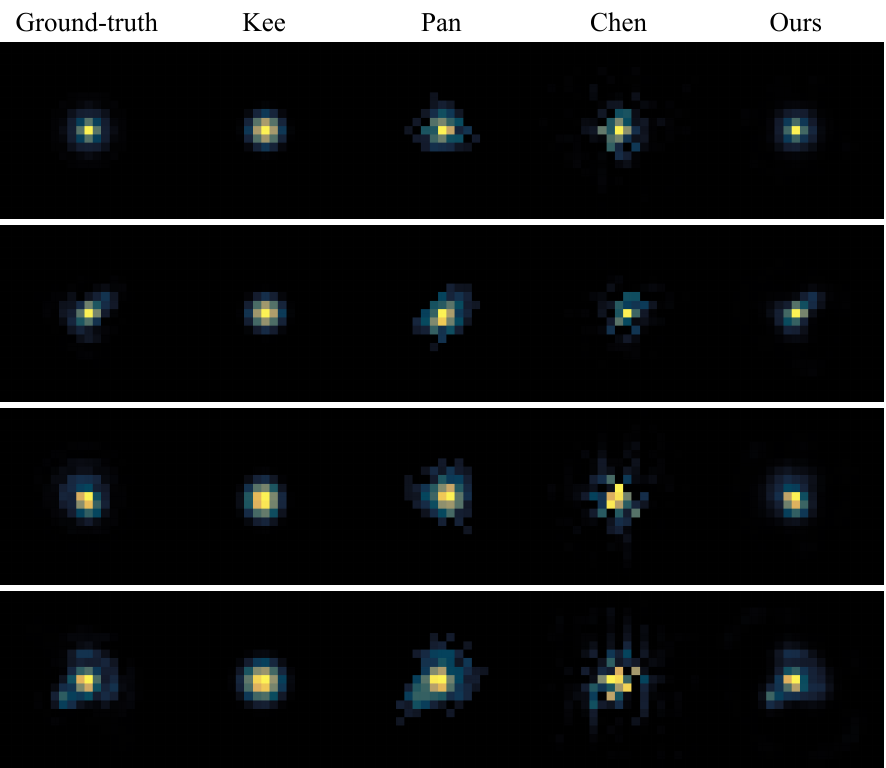}
    \caption{Comparison of estimated PSFs and ground truth at four field positions.}
    \label{fig:simulation results}
\end{figure}

\paragraph{Qualitative Results.}
As shown in \cref{fig:simulation results}, PSF estimation results from four representative field positions are visualized.  
\textbf{Kee et al.} produce over-concentrated kernels due to the Gaussian assumption.  
\textbf{Pan et al.} capture the main orientation but show noisy and truncated tails caused by sparsity-induced instability.  
\textbf{Chen et al.} exhibit fence-like artifacts due to checkerboard edges confined to two orientations.  
In contrast, \textbf{CircleFlow} reconstructs smooth PSFs with natural energy decay and consistent spatial structure, closely matching the ray-traced ground truth.  
The recovered kernels accurately preserve aberration signatures such as coma and astigmatism, demonstrating the method’s ability to recover spatially varying and locally anisotropic blur.

\begin{figure}[t]
    \centering
    \includegraphics[width=0.6\linewidth]{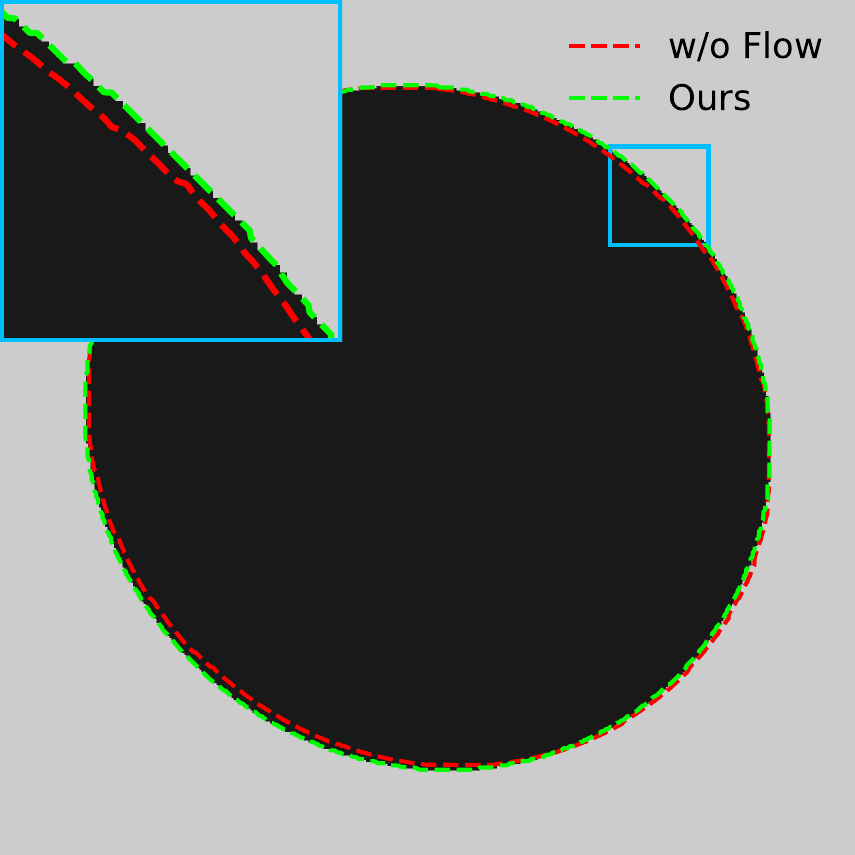}
    \caption{
    Edge overlay comparison in the ablation study.
    Reconstructed edges from CircleFlow (Ours) and the variant without flow guidance (w/o Flow) are overlaid on the ground-truth contour.
    }
    \label{fig:compare affine}
\end{figure}

\begin{figure*}[t]
    \centering
    \includegraphics[width=0.85\linewidth]{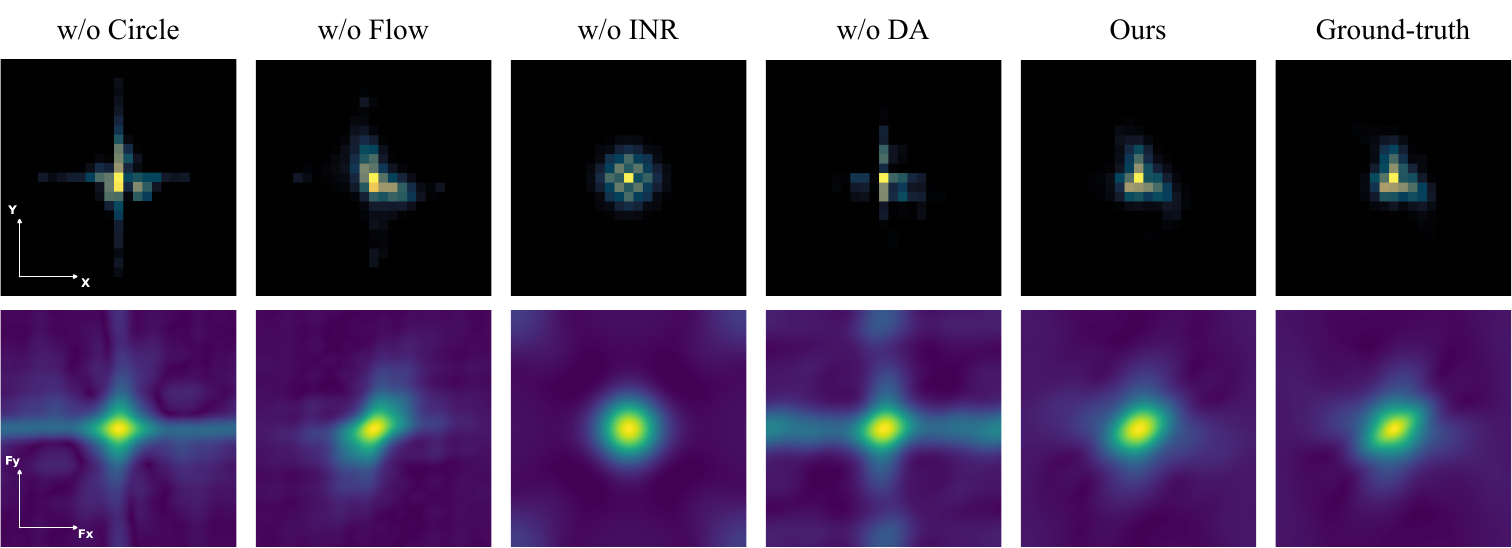}
    \caption{Ablation results showing PSFs and 2D MTFs under different configurations.}
    \label{fig:ablation}
\end{figure*}

\begin{figure*}[t]
    \centering
    \includegraphics[width=0.95\linewidth]{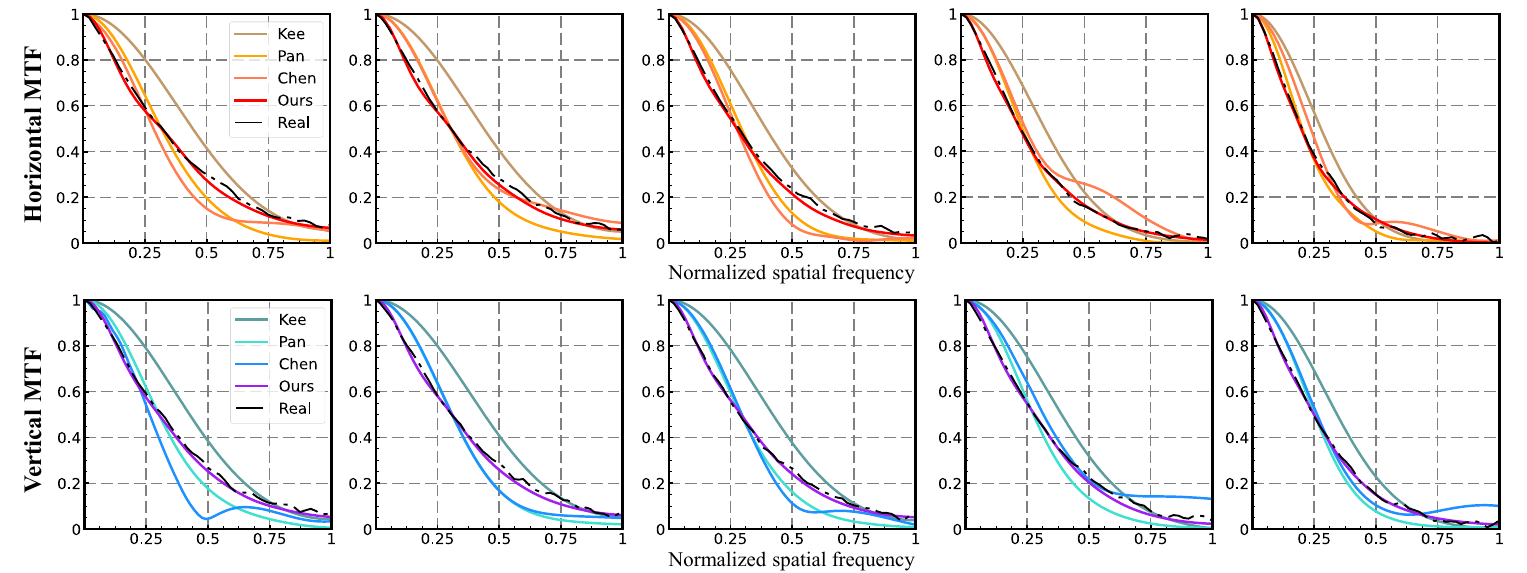}
    \caption{Real-world validation. Comparison of MTFs derived from estimated PSFs and the slanted-edge reference (Real).}
    \label{fig:sfr_measurement}
\end{figure*}

\begin{figure*}[t]
    \centering
    \includegraphics[width=\linewidth]{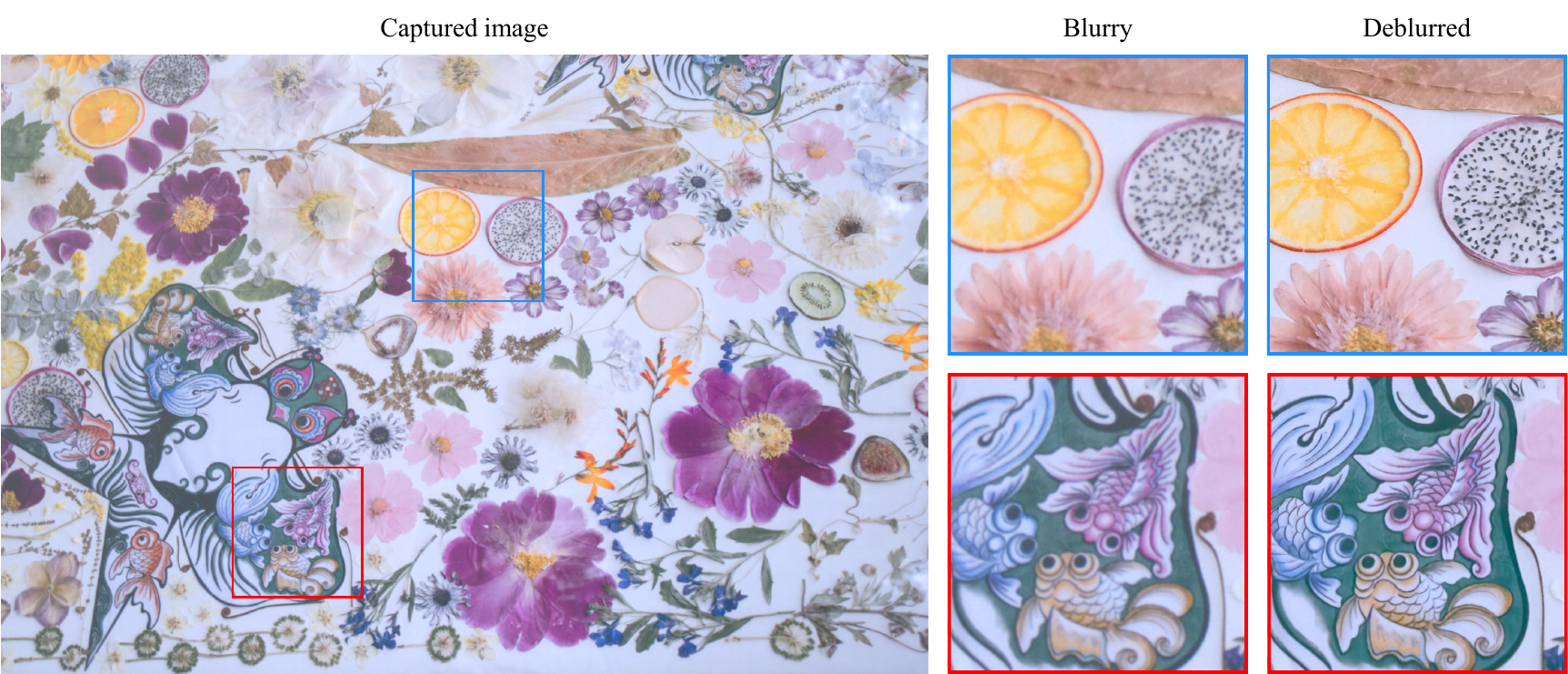}
    \caption{
    Deblurring results using PSFs estimated by CircleFlow.
    The left panel shows the captured blurred image, and the right panels display two representative regions before and after deblurring, where fine details are effectively restored.
    }
    \label{fig:deblur}
\end{figure*}

\begin{table}[t]
\vspace{-1mm}
\footnotesize
\caption{Quantitative comparison of PSF estimation on simulated data using the ray-traced smartphone telephoto lens model.}
\vspace{-2mm}
\label{tab:simulation_metrics}
\renewcommand{\arraystretch}{1.15}
\setlength{\tabcolsep}{12pt}
\centering
\resizebox{\linewidth}{!}{
\begin{tabular}{l|cccc}
\hline
 & Kee & Pan & Chen & Ours \\
\hline
PSNR (dB) & 32.87 & 30.08 & 28.96 & \textbf{43.91} \\
SSIM      & 0.913 & 0.892 & 0.871 & \textbf{0.982} \\
\hline
\end{tabular}
}
\vspace{-1.5mm}
\end{table}

\begin{table}[t]
\vspace{-1mm}
\footnotesize
\caption{Ablation study of CircleFlow on simulated data.}
\vspace{-2mm}
\label{tab:ablation}
\renewcommand{\arraystretch}{1.1}
\setlength{\tabcolsep}{3pt}
\centering
\resizebox{\linewidth}{!}{
\begin{tabular}{l|ccccc}
\hline
 & w/o Circle & w/o Flow & w/o INR & w/o DA & Ours \\
\hline
PSNR (dB) & 29.89 & 33.41 & 29.23 & 23.98 & \textbf{43.76} \\
SSIM      & 0.863 & 0.906 & 0.914 & 0.852 & \textbf{0.979} \\
\hline
\end{tabular}
}
\vspace{-1.5mm}
\end{table}

\paragraph{Quantitative Evaluation.}
\cref{tab:simulation_metrics} reports PSNR and SSIM between the estimated and ground-truth PSFs.  
CircleFlow achieves the highest scores in both metrics, outperforming competing approaches by a clear margin.  
These results confirm that CircleFlow delivers accurate PSF estimation under realistic imaging noise and sampling conditions.

\subsection{Ablation Studies}
\label{sec:ablation}

We further analyze the contribution of each key component in CircleFlow through ablation studies.  
Four controlled variants are designed for comparison:  
(1) \textbf{w/o Circle} – replacing the circle grid with a checkerboard pattern, causing directional bias and incomplete blur encoding;  
(2) \textbf{w/o Flow} – replacing optical flow with affine transformation estimated from corner detection, which fails to capture local deformation and results in edge misalignment, as shown in~\cref{fig:compare affine};  
(3) \textbf{w/o INR} – replacing the implicit neural PSF model with a Seidel basis expansion, limiting representation to low-order aberrations; and  
(4) \textbf{w/o DA} – disabling demosaicing awareness, which introduces interpolation artifacts that distort the recovered kernel.

\cref{fig:ablation} shows the estimated PSFs and corresponding 2D MTFs under different configurations, and \cref{tab:ablation} reports the quantitative results.
Together, they show that each component contributes meaningfully: the circle target encodes complete blur information, the flow guidance enables more precise, pixel-level edge localization, the INR models realistic energy distribution, and the demosaicing-aware formulation preserves optical fidelity.

\subsection{Real-World Experiments}
\label{sec:real_experiments}

We next validate CircleFlow on real camera data.  
A Sony $\alpha$6700 camera equipped with a Sony SELP1650 lens is used for data acquisition.  
A custom transmissive circle grid target illuminated by a D65 light source serves as the calibration chart, and all captures are recorded in RAW format to preserve sensor fidelity.  

For full-field characterization, the captured image is divided into $11\times17$ regions, yielding 187 PSFs per color channel, as shown in~\cref{fig:teaser}.
PSFs estimated from five representative field positions, spanning from the field center to the edge, are compared across methods.
The results show consistent trends with the simulations:  
\textbf{Kee et al.} produce over-concentrated kernels,  
\textbf{Pan et al.} yield noisy boundaries,  
\textbf{Chen et al.} show directional bias,  
while \textbf{CircleFlow} reconstructs spatially coherent PSFs with smooth energy decay and structural continuity across the field.

For quantitative validation, we conduct a spatial frequency response (SFR) test.
The MTFs derived from the estimated PSFs, corresponding to the same five field positions, are compared with reference MTFs obtained using the standard slanted-edge method.
As shown in \cref{fig:sfr_measurement}, CircleFlow’s MTFs align closely with the reference across all spatial frequencies, while other methods exhibit clear deviations.
This confirms that the recovered PSFs provide an accurate representation of real optical blur and are suitable for real-world calibration and lens evaluation.

\subsection{Deblurring with Recovered PSFs}
\label{sec:deblurring}

To further verify the practical utility of the recovered PSFs, we conduct a deblurring experiment on real photographs.  
Following the procedure in~\cite{chen2021optical}, a synthetic dataset is first generated by convolving sharp images with PSFs estimated by CircleFlow to train a learning-based deblurring network.  
The trained model is then applied to real blurred images captured with the same setup described in~\cref{sec:real_experiments}.  
As shown in~\cref{fig:deblur}, the reconstructed results exhibit clear improvement in fine texture restoration.  
These results not only demonstrate the applicability of the recovered PSFs in image restoration but also indirectly confirm the accuracy and reliability of the PSF estimation itself.
\section{Conclusion}
\label{sec:conclusion}

We presented CircleFlow, a high-fidelity PSF estimation framework that integrates structured circle-grid capture, a binary luminance prior, flow-guided edge localization, and an energy-constrained implicit neural PSF model.
This design transforms the ill-posed image–kernel coupling into a geometry-guided estimation problem, enabling subpixel-accurate edge reconstruction and stable recovery of spatially varying, anisotropic PSFs.
The demosaicing-aware joint optimization ensures that geometric alignment and blur modeling remain consistent with the sensor’s imaging pipeline.
Extensive simulation and real-world experiments show that CircleFlow produces PSFs matching ray-traced ground truth, aligns with measured SFR curves, and serves as an effective prior for image restoration.
We believe CircleFlow offers a practical path toward unified optical characterization and computational imaging enhancement, enabling tighter integration between optical calibration and data-driven processing.

{
    \small
    \bibliographystyle{ieeenat_fullname}
    \bibliography{main}
}

\clearpage

\setcounter{section}{0}
\setcounter{equation}{0}
\setcounter{table}{0}
\setcounter{figure}{0}
\setcounter{page}{1}

\maketitlesupplementary

\section*{Overview}

This supplementary document provides additional theoretical analysis and expanded experimental results that complement the main paper.

\begin{itemize}
    \item \textbf{Section~\ref{sec:binary_circle}} deepens the theoretical justification for using a binary circular boundary for PSF estimation, explaining how a circle provides a continuum of subpixel-shifted directional ESF measurements.
    \item \textbf{Section~\ref{sec:experiments}} presents additional experimental results, including:
    full-field PSF calibration across two imaging systems (\cref{fig:full_field}),
    challenging simulated PSF estimation scenarios (\cref{fig:sim}),
    and additional real-image deblurring results (\cref{fig:real_deblur}).
\end{itemize}

\section{Binary Edges, Circular Geometry, and Their Relation to ESF--LSF Theory}
\label{sec:binary_circle}

This section explains how binary edges and circular boundary geometry impose strong constraints on the blur formation model. These constraints make boundary profiles highly informative for estimating the PSF, especially when the edge is precisely localized.

\subsection{1D ESF--LSF relationship and the advantage of slanted-edge sampling}

In 1D imaging, the blur of a binary step edge is described by the edge spread function (ESF):
\begin{equation}
\mathrm{ESF}(x) = (H * k)(x),
\end{equation}
where $H(x)$ is the Heaviside step function and $k(x)$ is the 1D PSF.

Differentiation produces the line spread function (LSF):
\begin{equation}
\mathrm{LSF}(x) = \frac{d}{dx}\mathrm{ESF}(x)
                = (H * k)'(x)
                = H'(x)*k(x)
                \approx k(x).
\end{equation}

A classical refinement of this method is to capture the edge at a slight angle rather than perfectly aligned with the sampling grid. The oblique alignment causes the edge to intersect successive pixels at gradually shifted subpixel positions, effectively producing a dense set of ESF samples across the blur transition. This subpixel traversal mitigates sensitivity to the underlying sampling grid and leads to a better-resolved estimate of the blur profile.

Discretizing the 1D ESF--LSF relation yields a system
\begin{equation}
\mathbf{b} = H_{\mathrm{1D}}\mathbf{k},
\end{equation}
whose columns are shifted samples of the kernel. These columns remain strongly distinct, making $H_{\mathrm{1D}}$ typically close to full column rank.

\subsection{From 1D edges to 2D boundaries}

For a 2D blurred image $b = i*k$ and a boundary point $\mathbf{x}_0$, introduce local coordinates
\begin{equation}
u = (\mathbf{x}-\mathbf{x}_0)\!\cdot\!\mathbf{n}, \qquad
v = (\mathbf{x}-\mathbf{x}_0)\!\cdot\!\mathbf{t},
\end{equation}
where $\mathbf{n}$ and $\mathbf{t}$ denote the boundary normal and tangent.

A binary boundary behaves locally as a 1D step:
\begin{equation}
i(u,v) \approx H(u).
\end{equation}
Along the normal direction,
\begin{equation}
\partial_{\mathbf{n}} b(\mathbf{x}) 
= (\partial_{\mathbf{n}} i) * k
\approx \delta_{\mathrm{line}} * k,
\end{equation}
which mirrors the 1D ESF--LSF mechanism. A small boundary segment thus provides a directional ESF measurement, and discretization yields a system
\begin{equation}
\mathbf{b}_{\mathbf{n}} = H_{\mathbf{n}}\mathbf{k},
\end{equation}
whose structure again tends to exhibit strong column diversity.

\subsection{Circular boundary as a continuum of slanted-edge segments}

A binary circle 
\begin{equation}
i(\mathbf{x})= \mathbf{1}_{\|\mathbf{x}-\mathbf{c}\|\le R}
\end{equation}
yields boundary normals spanning all orientations $\theta\in[0,\pi)$.  
Each small arc on the circle forms a locally straight edge segment that intersects the sampling grid with a smoothly varying subpixel offset. Consequently, the circle implicitly provides a \emph{continuous sweep} of slanted-edge measurements: each oriented boundary segment samples the blur transition at slightly different subpixel phases, much like the 1D slanted-edge method but extended over all directions.

For each orientation $\theta$, the blurred boundary profile satisfies
\begin{equation}
b_{\theta}(u) = (H * k_{\theta})(u),
\end{equation}
leading to multiple linear systems
\begin{equation}
\mathbf{b}_\theta = H_{\theta}\mathbf{k}.
\end{equation}

Stacking them gives
\begin{equation}
\mathbf{B}=
\begin{bmatrix}
H_{\theta_1}\\
H_{\theta_2}\\
\vdots\\
H_{\theta_m}
\end{bmatrix}
\mathbf{k},
\end{equation}
where the combined operator
\begin{equation}
H_{\mathrm{circle}}=
\begin{bmatrix}
H_{\theta_1}\\
\vdots\\
H_{\theta_m}
\end{bmatrix}
\end{equation}
benefits from the circle's dense and smoothly varying subpixel sampling.  
This produces high column diversity and brings $H_{\mathrm{circle}}$ much closer to full column rank than operators derived from a single linear edge.

\subsection{Role of the binary prior}

The calibration target is binary:
\begin{equation}
i(\mathbf{x})\in\{0,1\},
\end{equation}
so smooth transitions around the boundary must originate from the PSF.  
Redistributing blur between $i$ and $k$ would break either the binary intensity pattern or the known circular geometry, preventing such transformations from matching the observed boundary shape.

\subsection{Summary}

Binary edges constrain blur to manifest explicitly in the captured image.  
A circle supplies a continuum of oriented, slanted-edge-like segments, providing dense and directionally diverse ESF measurements with rich subpixel coverage.  
The resulting linear operators exhibit strong column diversity and are typically close to full column rank, enabling accurate PSF estimation once the boundary is localized with high precision.

\section{Additional Experimental Results}
\label{sec:experiments}

\begin{figure}[h]
    \centering
    \includegraphics[width=\linewidth]{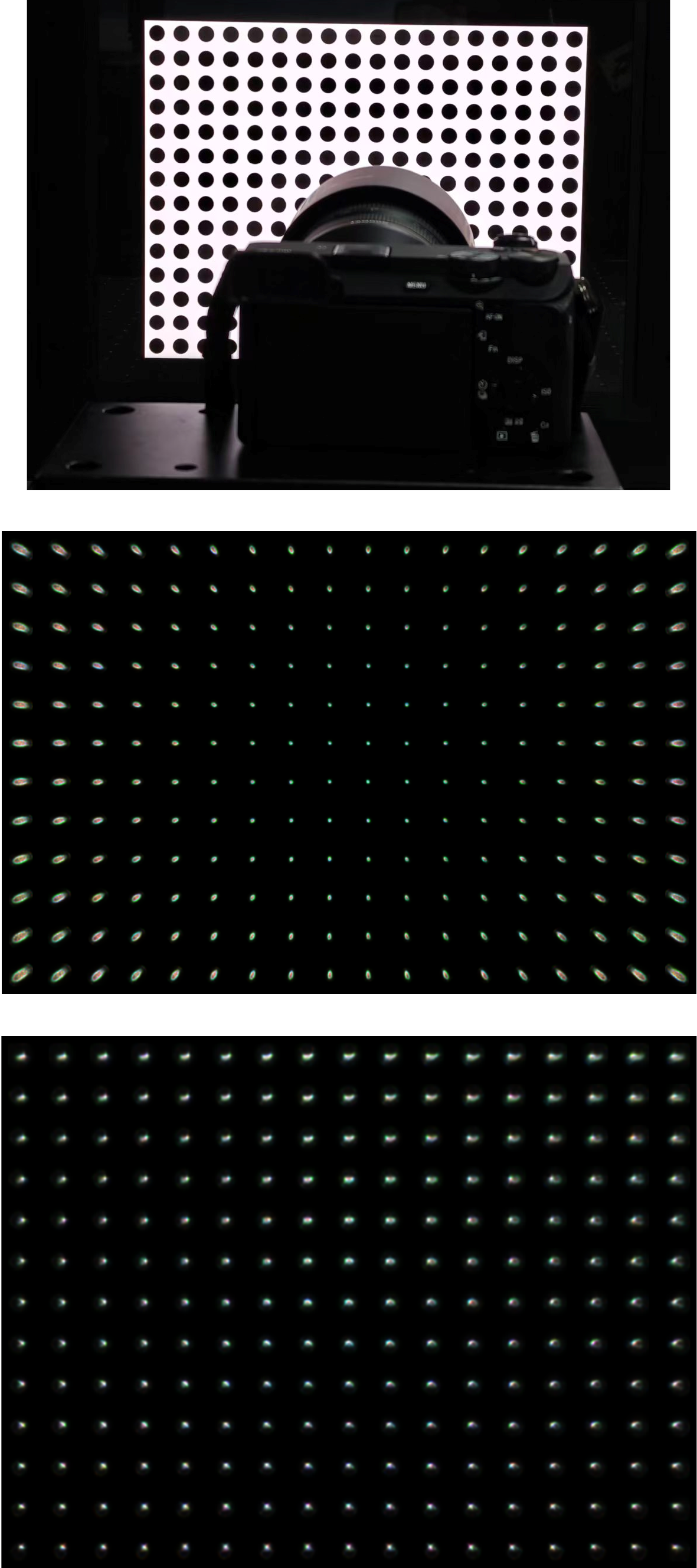}
    \caption{\textbf{Full-field PSF calibration on two imaging systems.}
    From top to bottom: 
    (1) experimental setup example using the Sony~$\alpha$6700 with the Tamron 70--300mm lens; 
    (2) full-field PSF calibration results for the Sony~$\alpha$6700 + Tamron 70--300mm lens; 
    (3) full-field PSF calibration results for the OnePlus~12 periscope telephoto module, whose PSFs lose their expected field-wise rotational symmetry due to lens-assembly tolerances. 
    CircleFlow consistently captures spatial PSF variations across both interchangeable-lens and mobile imaging systems.}
    \label{fig:full_field}
\end{figure}

\begin{figure*}[t]
    \centering
    \includegraphics[width=0.9\linewidth]{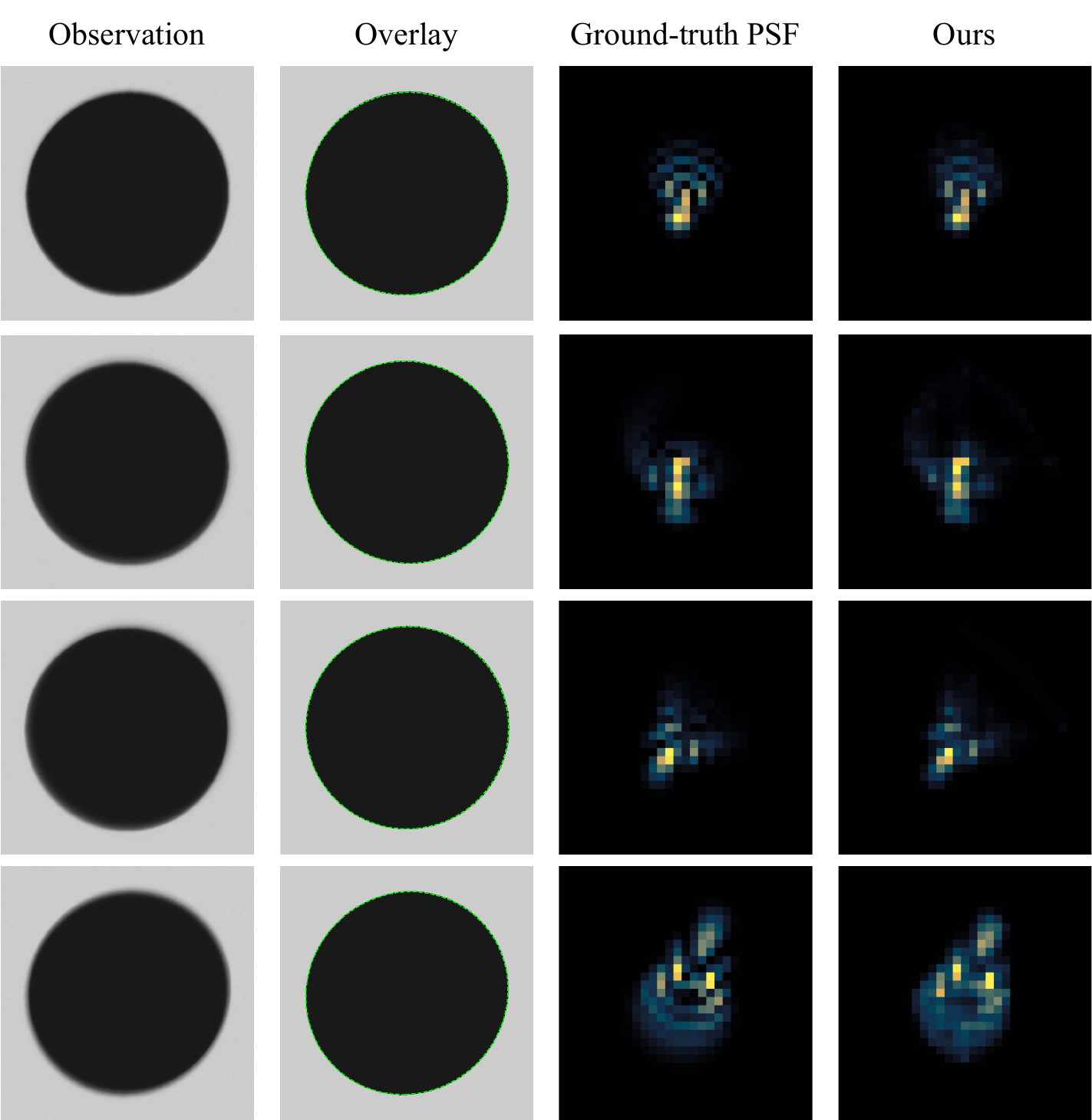}
    \caption{\textbf{PSF estimation under challenging simulation conditions.}
    From left to right: (1) blurred observations, 
    (2) recovered sharp edges overlaid on the ground-truth sharp images, 
    (3) ground-truth PSFs, 
    and (4) estimated PSFs. 
    CircleFlow reconstructs both the latent edge and the PSF with high fidelity.}
    \label{fig:sim}
\end{figure*}

\begin{figure*}[t]
    \centering
    \includegraphics[width=0.98\linewidth]{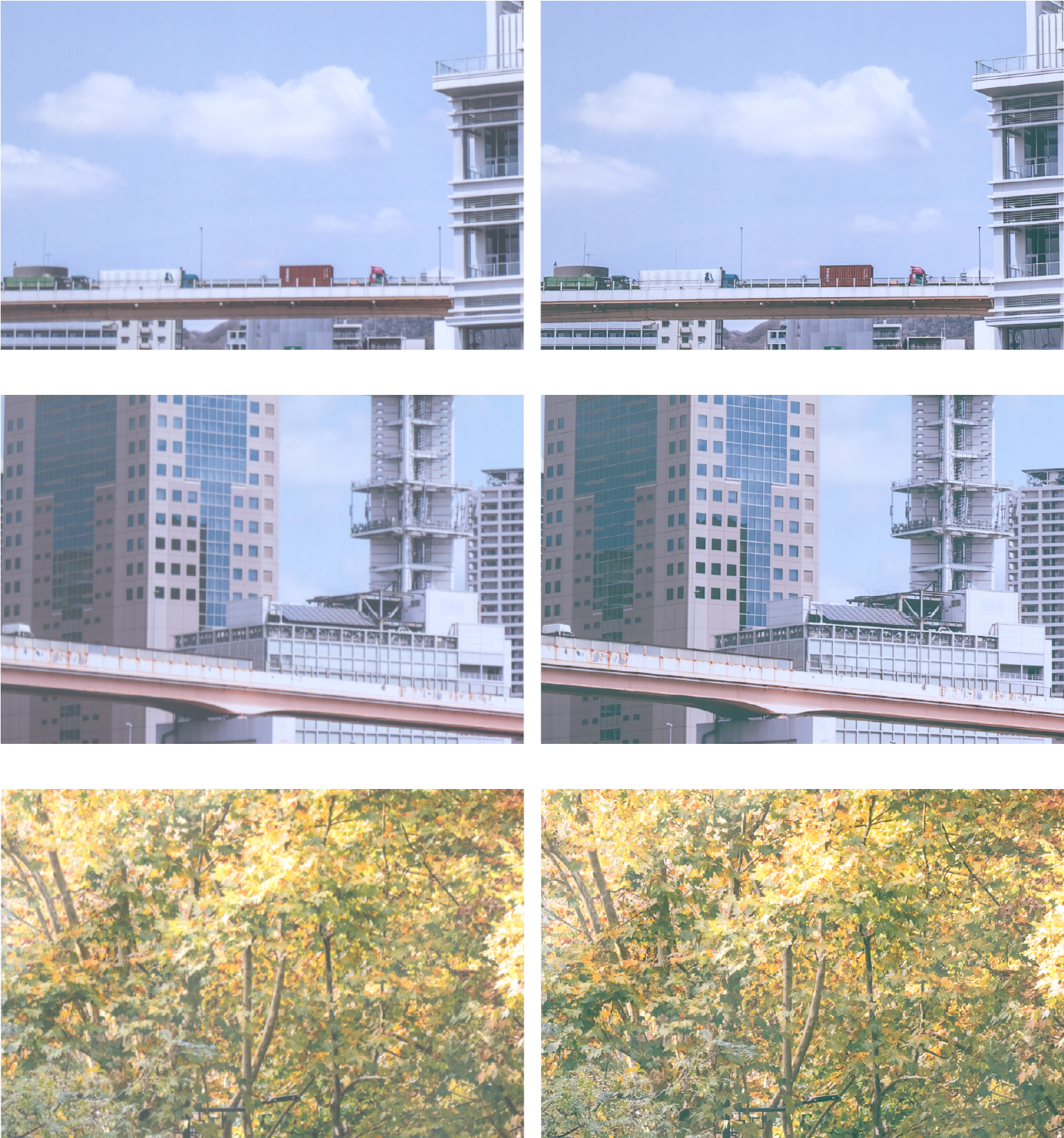}
    \caption{\textbf{Real-image deblurring results using CircleFlow-estimated PSFs.}
    The same deblurring pipeline described in the main paper is applied. 
    The results demonstrate that PSFs estimated from the circle grid target generalize well to real photographs,
    enabling faithful detail restoration.}
    \label{fig:real_deblur}
\end{figure*}

This section provides expanded empirical validation through three sets of additional results:
full-field PSF calibration on multiple imaging systems (\cref{fig:full_field}),
challenging simulated PSF estimation (\cref{fig:sim}),
and additional real-image deblurring examples (\cref{fig:real_deblur}).
These results further demonstrate the robustness and generality of CircleFlow.

\end{document}